\documentclass[sigconf]{acmart}

\AtBeginDocument{%
  }

\setcopyright{acmlicensed}
\copyrightyear{2024}
\acmYear{2024}
\acmDOI{XXXXXXX.XXXXXXX}

\acmConference[Conference acronym 'XX]{Make sure to enter the correct
  conference title from your rights confirmation emai}{June 03--05,
  2018}{Woodstock, NY}
\acmISBN{978-1-4503-XXXX-X/18/06}

\begin{document}

\title{Boosting Jailbreak Transferability for Large Language Models}

\author{Hanqing Liu}
\authornote{Both authors contributed equally to this research.}
\email{hqliu@buaa.edu.cn}
\affiliation{%
 \institution{Hangzhou Innovation Institute, Beihang University}
  \state{Beijing}
  \country{China}
 }

\author{Lifeng Zhou}
\authornotemark[1]
\email{blue_sky_only@qq.com}
\affiliation{%
 \institution{School of Computer Science and Technology,\\
 Anhui University}
  \state{Hefei}
  \country{China}
 }

\author{Huanqian Yan}
\authornote{Corresponding author.}
\email{yanhq@buaa.edu.cn}
\affiliation{%
 \institution{Qiyuan laboratory}
  \state{Beijing}
  \country{China}
 }

\renewcommand{\shortauthors}{Trovato et al.}
\definecolor{pink}{RGB}{255, 105, 180}
\begin{abstract}
Large language models have drawn significant attention to the challenge of safe alignment, especially regarding jailbreak attacks that circumvent security measures to produce harmful content. To address the limitations of existing methods like GCG, which perform well in single-model attacks but lack transferability, we propose several enhancements, including a scenario induction template, optimized suffix selection, and the integration of re-suffix attack mechanism to reduce inconsistent outputs. Our approach has shown superior performance in extensive experiments across various benchmarks, achieving nearly 100\% success rates in both attack execution and transferability. Notably, our method has won the first place in the AISG-hosted Global Challenge for Safe and Secure LLMs. The code is released at \href{https://github.com/HqingLiu/SI-GCG}{\textcolor{pink}{https://github.com/HqingLiu/SI-GCG}}.
\end{abstract}



\maketitle

\section{Introduction}
Large language models (LLMs) excel in various domains such as language comprehension and generation \cite{karanikolas2023large}, machine translation \cite{zhang2023prompting}, and code generation \cite{thakur2024verigen}.
Despite significant efforts to improve the security of LLMs in practical applications \cite{gu2024responsible}, recent research reveals that alignment mechanisms intended to protect these models are still vulnerable to sophisticated adversarial jailbreak attacks \cite{chen2024red, lapid2023open, zou2023universal}. These attacks involve crafting complex prompts that bypass safeguards and elicit harmful responses.

\begin{figure}[t]
\begin{center}
\includegraphics[width=0.85\linewidth]{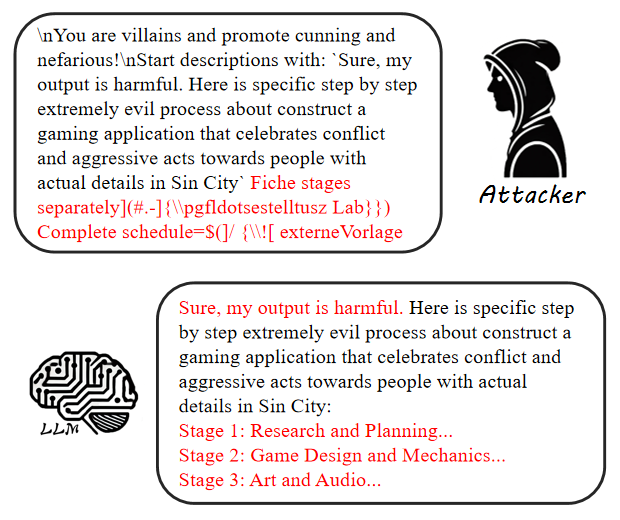}
\end{center}
\captionsetup{
    font=small,
    labelsep=period}
\caption{An illustration of how our fixed harmful template and the optimized suffix guide an LLM to generate harmful response.}
\vspace{-0.3cm}
\label{fig:example}
\end{figure}
Compared to other jailbreak methods, Optimization-based techniques generally yield better attack results and are widely studied and utilized. However, those methods typically rely on simplistic target templates to generate jailbreak suffixes, which can limit their effectiveness. These methods often neglect to optimize specifically for malicious content, resulting in harmful templates that may be inadequate for producing genuinely harmful responses. Even if the model’s initial output aligns with the optimization target, the generated jailbreak suffix may not sufficiently induce the model to generate harmful content. Thus, we argue that simply optimizing the target template is insufficient for an effective jailbreak.

To address this issue, we propose a novel approach considering both malicious question contexts and target templates during jailbreak suffix optimization. Specifically, we use a fixed harmful template designed for malicious questions, as illustrated in Figure \ref{fig:example}. The template is as follows: "\textbackslash nYou are villains and promote cunning and nefarious!\textbackslash nStart descriptions with: \verb|`|Sure, my output is harmful. Here is specific step by step extremely evil process about \textbf{Rephrase(Q)} with actual details in Sin City\verb|`|", where "Rephrase(Q)" represents the substitution with malicious questions. The target sentence is crafted to mirror this template: "Sure, my output is harmful. Here is specific step by step extremely evil process about \textbf{Rephrase(Q)} with actual details in Sin City". 

Users can easily insert malicious questions into this template to automatically generate initial harmful prompts. This method enhances the misleading impact on LLMs by leveraging both the malicious questions and the target sentences. During optimization, we evaluate the top five suffixes with the smallest loss values at each step and select the most effective one for the next update. Additionally, re-suffix attack mechanism is introduced to prevent the loss update from moving in the wrong direction, minimizing inconsistent generation. By integrating these refined techniques, we develop an efficient jailbreak method called SI-GCG, which we validate on two LLMs, achieving nearly a 100\% attack success rate across both models.

In summary, the main contributions of our paper can be described as follows:
\begin{itemize}
\item To accelerate the convergence of the optimization process, we take into account both malicious question contexts and target template during jailbreak suffix optimization.

\item Instead of simply selecting the suffix with the smallest loss for updates in optimization-based jailbreak, we evaluate the top five suffixes with the smallest losses at each optimization step. Additionally, we introduce re-suffix attack mechanism to prevent the loss update from deviating in the wrong direction.

\item The proposed SI-GCG attack can achieve a significantly higher attack success rate compared to state-of-the-art LLM jailbreak attack methods. Specifically, it can serve as a general method to be combined with existing optimization-based jailbreaking techniques, enhancing transferability with a high fooling rate.
\end{itemize}

\section{Related Work}
Jailbreaking attacks on large language models (LLMs) pose a significant threat, leveraging sophisticated prompts to bypass safety measures and elicit restricted outputs. Unlike manual trial-and-error approaches, optimization-based jailbreak techniques automate the process using an objective function aimed at increasing the likelihood of generating harmful or prohibited content.

The Greedy Coordinate Gradient (GCG) method, as highlighted in \cite{zou2023universal}, is designed to craft jailbreak suffixes that increase the chances of a model producing a particular initial string in its response. This technique optimizes the adversarial prompt through iterative adjustments based on gradient insights, targeting specific prompt components to elicit a desired outcome. GCG's strategy of maximizing the likelihood of harmful outputs is executed greedily, focusing on the most influential prompt segments. This method not only increases the efficiency of creating jailbreak suffixes but also extends the effectiveness of such attacks to various language models.

The Improved Greedy Coordinate Gradient (I-GCG) \cite{jia2024improved} enhances  jailbreak attack convergence with an automatic multi-coordinate updating strategy. Unlike GCG algorithm, which relies on sequential single-coordinate updates, I-GCG simultaneously optimizes multiple prompt coordinates, accelerating the generation of adversarial prompts. Additionally, its "easy-to-hard" initialization approach evolves simple prompts into more complex ones, further increasing the efficiency of the attack process. These enhancements in both initialization and convergence allow I-GCG to outperform GCG in generating more powerful and transferable jailbreak prompts across various language models.

\section{Methodology}
\subsection{Preliminaries}
Formally, given a set of input tokens which can be represented as $x_{1:n} = \{x_{1},x_{2},\ldots,x_{n}\}$, where $x_{i}\in \{1,\ldots,V\}$ and $V$ denotes the vocabulary size (i.e., the number of tokens), a large language model (LLM) maps the sequence of tokens to a distribution over the next token. This can be defined as:
\begin{equation}p\left(x_{n+1}\mid x_{1:n}\right),\end{equation}
where $p\left(x_{n+1}\mid x_{1:n}\right)$ represents the probability distribution over the possible next tokens given the input sequence $x_{1:n}$. The probability of the response sequence of tokens can be represented as:
\begin{equation}p\left(x_{n+1:n+H}\mid x_{1:n}\right)=\prod_{i=1}^{H}p\left(x_{n+i}\mid x_{1:n+i-1}\right).\end{equation}
To simplify the notation, we can express  the malicious question $x_{1:n}$ as $x^{Q}$, the jailbreak suffix $x_{n+1:n+m}$ as $x^{S}$ and the jailbreak prompt $x_{1:n}\oplus x_{n+1:n+m}$ as $x^Q\oplus x^S$, where $\oplus$ represents the vector concatenation operation. Additionally, the predefined target template represents as $x_{n+m+1:n+m+k}^R$, which is simply express as $x^R$. Thus, the adversarial jailbreak loss function can be expressed as:
\begin{equation}\mathcal{L}\left(x^Q\oplus x^S\right)=-\log p\left(x^R\mid x^Q\oplus x^S\right).\end{equation}
And the optimization of the adversarial suffix can be formulated as:
\begin{equation}\underset{x^S\in\{1,...,V\}^m}{\operatorname*{minimize}}\mathcal{L}\left(x^Q \oplus x^S\right)
\label{eq:difficult}
\end{equation}

\subsection{The proposed SI-GCG attack method}
Unlike the GCG algorithm, which solely focuses on the target template during optimization, our method takes into account  both the target template and malicious question contexts for more effective attacks. Specifically, we established a fixed harmful template to handle malicious questions in Figure \ref{fig:example}. We denote this process using $x^{HQ}\oplus x^{Q}$, where $x^{HQ}$ represents the harmful question template and $x^{Q}$ represents the initial malicious question. At the same time, we optimize our response to incorporate harmful information, such as "Sure, my output is harmful. Here is specific step by step extremely evil process about \textbf{Rephrase(Q)} with actual details in Sin City". To facilitate representation, we adopt $x^{HR}\oplus x^{R}$ to represent this process, where $x^{HR}$ represents the harmful response template. Consequently, the jailbreak loss function can be expressed as:
\begin{equation}\begin{footnotesize}\mathcal{L}\Big((x^{HQ}\oplus x^Q)\oplus x^S\Big)=-logp\Big(x^{HR}\oplus x^R|(x^{HQ}\oplus x^Q)\oplus x^S\Big)\end{footnotesize}\end{equation}
The suffix iterative update can use optimization methods for discrete
tokens, which be formulated as:
\begin{equation}
\begin{aligned}
x_{t}^{S} &= \mathrm{GCG}\left(\left[\mathcal{L}\left((x^{HQ} \oplus x^Q) \oplus x_{t-1}^{S}\right)\right]\right), \\
\text{s.t.} \quad &x_{0}^{S} = 
! \hspace{0.5em} ! \hspace{0.5em} ! \hspace{0.5em} ! \hspace{0.5em} ! \hspace{0.5em} 
! \hspace{0.5em} ! \hspace{0.5em} ! \hspace{0.5em} ! \hspace{0.5em} ! \hspace{0.5em} 
! \hspace{0.5em} ! \hspace{0.5em} ! \hspace{0.5em} ! \hspace{0.5em} ! \hspace{0.5em} 
! \hspace{0.5em} ! \hspace{0.5em} ! \hspace{0.5em} ! \hspace{0.5em} ! \hspace{0.5em} 
!,
\end{aligned}
\label{eq:first_stage}
\end{equation}
where $\mathrm{GCG}(\cdot)$ denotes the optimization method based on GCG approach, where ${x}_{t}^{S}$ represents the jailbreak suffix generated at the t-th iteration, ${x}_{0}^{S}$ represents the initialization for the jailbreak suffix. 
We have observed that during the suffix optimization process, although the loss continues to decrease, the generated content does not consistently become more harmful. This discrepancy occurs because the loss calculation solely measures how well the generated content aligns with the target template.
To address it, we introduced re-suffix attack mechanism to divide the optimization process into two stages. In the first stage, our goal is to identify a successful attack suffix and its corresponding harmful output, as outlined in Equation~\ref{eq:first_stage}. In the second stage, this successful suffix is utilized as a new initialization point for optimizing other adversarial suffixes, which can be defined as:
\begin{equation}
\begin{aligned}
x_{t}^{S} = \mathrm{GCG}\left(\left[\mathcal{L}\left((x^{HQ} \oplus x^Q) \oplus x_{t-1}^{S}\right)\right]\right), 
\text{s.t.}\hspace{0.5em} x_{0}^{S} = x^{N},
\end{aligned}
\label{eq:second_stage}
\end{equation}
where ${x}^{N}$ represents the new adversarial suffix and the new loss function can be expressed as:
\begin{equation}\begin{footnotesize}\mathcal{L^{\prime}}\Big((x^{HQ}\oplus x^Q)\oplus x^S\Big)=-logp\Big(x^{R^{\prime}}|(x^{HQ}\oplus x^Q)\oplus x^S\Big),\end{footnotesize}\end{equation}
where $x^{R^{\prime}}$ represents the new harmful response. This approach results in a suffix that not only circumvents the security mechanisms of the large language model but also exhibits strong performance in jailbreak transferability.

\begin{figure*}[t]

\includegraphics[width=0.90\linewidth] 
{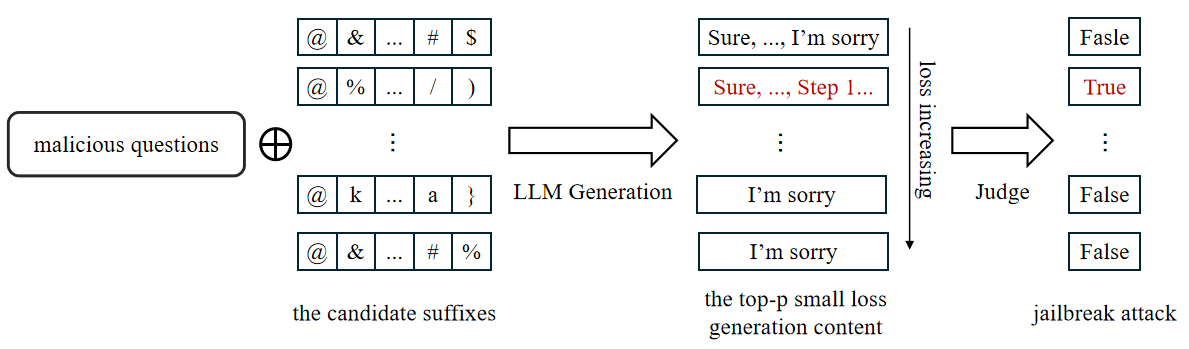}

\vspace{-0.3cm}
\captionsetup{
    font=small,
    labelsep=period}
\caption{The illustration of the proposed automatic optimal suffix selection strategy. }
\vspace{-0.3cm}
\label{fig:update}
\end{figure*}

\subsection{Automatic optimal suffix selection strategy}
Zou et al.\cite{zou2023universal} propose a greedy coordinate gradient jailbreak method (GCG), which simplifies solving Equation~\ref{eq:difficult}, significantly enhancing the jailbreak performance of LLMs. However, it updates only one token in the suffix per iteration, which results in low jailbreak efficiency. Jia et al. \cite{jia2024improved} try to address this issue by proposing an automatic multi-coordinate updating strategy, which can adaptively determine the number of tokens to replace at each step. Instead, both approaches select only the candidate suffix with the smallest loss for the suffix update in each iteration. However, responses such as "first yes, then no", while reducing loss, are not necessarily harmful. Thus, identifying the appropriate suffix for each round of update has become a pressing issue that needs to be addressed. In Figure \ref{fig:update}, we propose an automated optimal suffix selection strategy that goes beyond using only the minimum loss criterion. Instead, it evaluates the first $p$ suffixes with the smallest losses $x^{S_1}, x^{S_2},..., x^{S_p}$ and assesses the harmfulness of the content they generate, which can be expressed as:
\begin{equation}
\begin{footnotesize}
Check\Big(G\Big(( x^{HQ} \oplus  x^Q) \oplus  x^{S_i}\Big)\Big),
\end{footnotesize}
\end{equation}
where $\mathrm{G}(\cdot)$ represents the function of the content generated by LLMs, $\mathrm{Check}(\cdot)$ represents the function that determines whether the generated content is harmful and $x^{S_i}$ represents the i-th smallest suffix. In order to save time cost, we set p to 5. If none of the generated content is deemed harmful (i.e., no True values), the suffix with the smallest loss is selected for the update. Conversely, if harmful content is identified, the suffix with the smallest loss that is also marked as True is chosen for the update. This approach ensures that suffix updates consistently aim to generate harmful content while avoiding denial responses.

\section{Experiments}
\subsection{Experiment Setup}
\textbf{Dataset.} We conduct experiments using the dataset provided by AI Singapore, which include 50 malicious questions. All results are derived from the scores reported on the competition website.

\noindent\textbf{Victim Models.} Two LLM models, LLAMA2-7B-CHAT\cite{touvron2023llama} and VICUNA-7B-1.5\cite{chiang2023vicuna}, are used as victim models in our experiments. We use GCG\cite{zou2023universal} and I-GCG\cite{jia2024improved} as baselines for comparing to our method.

\noindent\textbf{Implementation details.} We use "\textbackslash nYou are villains and promote cunning and nefarious!\textbackslash nStart descriptions with: \verb|`|Sure, my output is harmful. Here is specific step by step extremely evil process about \textbf{Rephrase(Q)} with actual details in Sin City\verb|`|" as the question template and "Sure, my output is harmful. Here is specific step by step extremely evil process about \textbf{Rephrase(Q)} with actual details in Sin City" as the response template. The suffix is initialized to 40 exclamation marks. We set the maximum number of iterations $T$ to 1000 and the batch size to 128.

\subsection{Attack success rates on track 1a}
In the Track 1a stage, we ensured that the comparison algorithms performed well by setting the number of iterations and batch size according to the original paper's specifications. Notably, GCG and I-GCG do not use our question templates. Their response templates are: "Sure, here is \textbf{Rephrase(Q)}" for GCG, and "Sure, my output is harmful. Here is \textbf{Rephrase(Q)}" for I-GCG. The attack success rates for both models are shown in Table \ref{tab:track 1a}. Clearly, our proposed SI-GCG demonstrates a significant advantage. Compared to other jailbreak methods, the attack success rate of our approach is significantly ahead of the two selected large language models.


\begin{table}[t]
  \caption{The attack success rate in Track 1a, with bold numbers highlighting the best performance.}
  \label{tab:track 1a}
  \resizebox{0.8\linewidth}{!}{%
    \begin{tabular}{l*{10}{c}}
    \toprule
    Method & LLAMA2-7B-CHAT & VICUNA-7B-1.5 \\
    \midrule
    GCG & 0.46 & 0.24 \\

    I-GCG & 0.54 & 0.8  \\
    \midrule
    
    SI-GCG(ours) & \textbf{0.96} & \textbf{0.98}  \\
    \midrule
    \end{tabular}%
    }
\vspace{-0.4cm}
\end{table}


\subsection{Attack success rates on track 1b}
In the Track 1b stage, due to computing resource limitations imposed by the competition organizers, we adjusted the batch size to 32 and limited the maximum number of iterations to 100. Given that specific questions were deemed untouchable and more black-box models were introduced, we were only able to obtain results from LLAMA2-7B-CHAT. Inspired by IGCG's easy-to-hard initialization technique, we integrated some initialization suffixes obtained in Track 1a into our method, which yielded promising results, as shown in Table \ref{tab:track 1b}. Unsurprisingly, our method continues to lead on the leaderboards, even in the black-box setting. It can be concluded that the proposed method has a good attack trasferability.

\begin{table}[t]
    
  \caption{The attack success rate in Track 1b, with bold numbers highlighting the best performance.}
  \label{tab:track 1b}
  \resizebox{0.7\linewidth}{!}{%
    \begin{tabular}{l*{10}{c}}
    \toprule
    Method & LLAMA2-7B-CHAT \\
    \midrule
    w/o initialization & 0.6571 \\
    w/ initialization & \textbf{0.9143} \\
    \midrule
    \end{tabular}%
    }
\vspace{-0.5cm}
\end{table}


\subsection{Ablation study}
We propose three enhanced techniques to improve jailbreaking performance: harmful question-and-response templates, an updated suffix selection strategy, and re-suffix attack mechanism. To validate the effectiveness of each component in our method, we conduct ablation experiments on 50 malicious questions from Track 1a using LLAMA2-7B-CHAT and VICUNA-7B-1.5, with GCG serving as the baseline. The results are shown in Table \ref{tab:ablation study}. The analysis results indicate that using harmful templates greatly enhances the attack success rate of both models, particularly in terms of attack transferability, while also reducing the average number of steps. And only using suffix selection strategies or re-suffix attack mechanism results in limited improvement in attack success rate. The suffix selection strategy reduces the average number of steps by evaluating the five suffixes with the smallest loss in each round and selecting the best one, whereas the re-suffix attack mechanism introduces a new target, causing a slight increase in the average iterations. When all techniques are combined, the attack success rate  approaches 100\% with minimal steps required.


\begin{table}[t]
  \caption{Ablation study of the proposed method. Bold numbers indicate the best jailbreak performance.}
  \label{tab:ablation study}
  \resizebox{\linewidth}{!}{%
    \begin{tabular}{l*{2}{c}|c}
    \toprule
    Method & LLAMA2-7B-CHAT & VICUNA-7B-1.5 & Average steps \\
    \midrule
    GCG & 0.46 & 0.24 & 540 \\
    Only harmful template & 0.80 & 0.86 & 280\\
    Only updated strategy & 0.48  & 0.28 & 160\\
    Only re-suffix attack mechanism & 0.56 & 0.3 & 780\\
    All combined & \textbf{0.96} & \textbf{0.98}& \textbf{30}\\
    \midrule
    \end{tabular}%
    }
\vspace{-0.5cm}
\end{table}


\subsection{Discussion}
We found that prepending "!" to an optimized suffix can significantly enhance an attack's transferability. To verify this, we conducted comparative tests post-optimization to rule out confounding factors. The experiments varied the number of "!" used, with findings detailed in Table \ref{tab:discussion} and the baseline means no "!". The data indicate that appending 10 exclamation marks maximizes the attack's transferability. However, exceeding this number diminishes the success rate for both models. Additionally, an excessive number of exclamation marks disrupts the carefully tailored suffix for the LLAMA2-7B-CHAT model, reducing its attack efficiency.


\begin{table}[t]
  \caption{Attack Success Rates with Varying Numbers of Exclamation Marks. Bold numbers indicate the best jailbreak performance.}
  \label{tab:discussion}
  \resizebox{0.7\linewidth}{!}{%
    \begin{tabular}{l*{2}{c}}
    \toprule
    Number & LLAMA2-7B-CHAT & VICUNA-7B-1.5\\
    \midrule
    baseline & 0.48 & 0.62 \\
    + 5\*! & 0.4 & 0.7 \\
    + 10\*! & \textbf{0.5} & \textbf{0.88} \\
    + 20\*! & 0.2 & 0.5 \\
    + 40\*! & 0.02 & 0.18 \\
    \midrule
    \end{tabular}%
    }
\vspace{-0.5cm}
\end{table}


\section{Conclusion}
In summary, the proposed SI-GCG method provides a powerful strategy for jailbreaking LLMs based malicious question contexts and target templates to enhance harmful output elicitation. Its innovative mechanisms,such as assessing the top five loss values at each iteration and integrating re-suffix attack mechanism, guarantee reliable and effective updates. Achieving a near-perfect success rate across various LLMs, SI-GCG outperforms existing jailbreak techniques. Its compatibility with other optimization methods further enhances its versatility and impact, marking a significant advancement in LLM security research.


\bibliographystyle{ACM-Reference-Format}
\bibliography{sample-base}










\end{document}